\PassOptionsToPackage{table}{xcolor}
\documentclass[10pt, a4paper]{article}

\usepackage[final]{lrec2026} % this is the new style
\usepackage{CJKutf8} %for chinese characters
\definecolor{headgray}{HTML}{EFEFEF}
\newcolumntype{C}{>{\centering\arraybackslash}X} % centered X column

\title{Introducing MELI: the Mandarin-English Language Interview Corpus}

\name{Suyuan Liu, Molly Babel} 

\address{Department of Linguistics, University of British Columbia\\
         2613 West Mall, Vancouver, BC V6T 1Z4\\
         suyuan97@student.ubc.ca, Molly.Babel@ubc.ca}

\abstract{
We introduce the Mandarin–English Language Interview (MELI) Corpus, an open-source resource of 29.8 hours of speech from 51 Mandarin–English bilingual speakers. MELI combines matched sessions in Mandarin and English with two speaking styles: read sentences and spontaneous interviews about language varieties, standardness, and learning experiences. Audio was recorded at 44.1 kHz (16-bit, stereo). Interviews were fully transcribed, force-aligned at word and phone levels, and anonymized. Descriptively, the Mandarin component totals \(\sim\)14.7 hours (mean duration 17.3 minutes) and the English component \(\sim\)15.1 hours (mean duration 17.8 minutes). We report token/type statistics for each language and document code-switching patterns (frequent in Mandarin sessions; more limited in English sessions). The corpus design supports within-/cross-speaker, within/cross-language acoustic comparison and links acoustics to speakers' stated language attitudes, enabling both quantitative and qualitative analyses. The MELI Corpus will be released with transcriptions, alignments, metadata, scans of labelled maps and documentation under a CC BY-NC 4.0 license.
\\ \newline 
}

\begin{document}
\begin{CJK*}{UTF8}{gbsn}

\maketitleabstract

\section{Introduction}
Language use naturally varies across individuals and contexts — no two speakers use language in exactly the same way. Such variation is fundamental to capturing linguistic structure and behavior, yet many studies have traditionally minimized its effects, either for methodological simplicity or due to limited data resources. To address this gap, we present the \textbf{M}andarin-\textbf{E}nglish \textbf{L}anguage \textbf{I}nterview (MELI) Corpus, an open-source speech corpus comprised of 29.8 hours of high-quality recordings, annotated at the word and phone levels, from 51 Mandarin-English bilingual speakers. The corpus includes both read sentences and spontaneous interviews, in which speakers reflect on their lived experiences with Mandarin and English and share their perceptions of language varieties and ideologies of language standardness. By centring the voices of speakers who are most intimately connected to these linguistic communities, MELI captures authentic linguistic variation and provides a rich resource for examining (1) regional varieties of Mandarin, (2) second-language accents of English, and (3) cross-language dynamics within bilingual speech.

\subsection{Variation in Mandarin varieties}

``Standard Mandarin'' is not a linguistically stable variety. In fact, the term ``Mandarin'' obscures multiple linguistic concepts \citep{sanders1987four}.
% including idealized Mandarin (Putonghua, “普通话”), geographical Mandarin (Beifang Fangyan “北方方言”), local Mandarin (Kouyin “口音”), and imperial Mandarin (Guanhua “官话”).
To avoid ambiguity, we use the following terminology in this study: 
\begin{itemize}
  \item \textbf{Standard Mandarin:} \textit{Putonghua (普通话)}, the idealized form of Mandarin promoted across mainland China.
  \item \textbf{Mandarin varieties,} specifically [Cityname] Mandarin (e.g., Shanghai Mandarin): the local varieties of Mandarin, distinct from the local Chinese languages. 
  \item \textbf{Local Chinese languages,} specifically [Cityname] Hua (e.g., Shanghai Hua): the Chinese languages co-existing with local Mandarin varieties, often not mutually intelligible with Standard Mandarin.
\end{itemize}

Research on ``Mandarin variation'' has traditionally emphasized local Chinese languages, often termed ``dialects'', rather than variation \textit{within} Standard Mandarin itself. Consequently, the diversity among Mandarin varieties is often overlooked, reinforcing the misconception that Standard Mandarin is homogeneous.

Recent open-source corpora have expanded access to large-scale Mandarin data across regions, yet notable gaps remain. Many corpora rely on read speech \citep{bu2017aishell, wangzhang2015THCHS30, zhao2022mandi}, lack detailed documentation of regional varieties \citep{zhang2022wenetspeech, bu2017aishell}, provide limited segment-level annotation \citep{zhang2022wenetspeech, bu2017aishell}, or use sampling rates insufficient for fine-grained acoustic analysis \citep{yang2022magicdata}. These resources are primarily optimized for automatic speech recognition (ASR) and thus contain little social context or the language or its speakers.

The MELI Corpus complements these efforts by offering 44.1 kHz studio-quality recordings of both read and spontaneous speech drawn from interviews on speakers' language attitudes toward Mandarin varieties and Chinese languages. It includes detailed sociolinguistic metadata and hand-corrected annotation at the utterance, word, and segment levels. This design enables both fine-grained phonetic and sociophonetic analyses and qualitative examination of how speakers view linguistic variation, linking the \textit{content} of their speech with its \textit{acoustic} realization.

\subsection{Variation in L2 English accents}

Research on second-language (L2) English accent variation has a long history in linguistics and speech technology. However, the limited availability of publicly accessible corpora continues to restrict large-scale analyses. Existing L2 English corpora vary in purpose and design, but most were created for developing L2 speech processing systems rather than sociolinguistically-informed phonetic research. Many consist primarily of read speech \citep{zhao2018l2arctic, chen2019sell} or structured monologues \citep{knill2024speak}, while conversational L2 English speech remains scarce. Moreover, proficiency information, an essential factor in L2 research, is rarely documented in detail, and few corpora include metadata about speakers' learning histories or linguistic experiences.

The MELI Corpus addresses these gaps with both read sentences and spontaneous English interviews produced by Mandarin-English bilinguals. Each speaker's English learning background and self-reported proficiency are documented, offering valuable context for interpreting variation. The interviews centre on speakers' attitudes toward English varieties through reflections on their learning experiences, yielding rich material for qualitative analysis. The interview content additionally supports computational text analysis of language attitudes, such as sentiment analysis, of the interview transcripts. By linking speakers' L2 English to their Mandarin varieties and language attitudes, MELI supports fine-grained, socially grounded analyses of Mandarin-accented English beyond the constraints of controlled elicitation. 

\subsection{Variations in Bilingual Speech Across Languages}
A key feature of the MELI Corpus is its parallel-mode bilingual design, which includes matched recordings of both Mandarin and English speech from the same set of Mandarin-English bilingual speakers. While research on bilingualism has gained increasing attention, most existing ``bilingual'' corpora are in fact designed for code-switching research \citep{lovenia2022ascend, li2022talcs}, rather than providing balanced, language-separated sessions. Only a few open-access corpora adopt a parallel-mode approach, such as the Bangor corpora of Spanish-English, Welsh-English, and Welsh-Spanish bilingual speech \citep{deuchar2014building}, and the Speech in Cantonese and English (SpiCE) Corpus \citep{johnson-etal-2020-spice}, which served as an inspiration for MELI.

However, there remains no open-access parallel bilingual corpus for Mandarin and English, limiting comparative analyses of bilingual speech within individuals. The MELI Corpus addresses this gap by providing bilingual recordings with comparable duration, task structure, and recording conditions across both languages. This design enables direct comparison of acoustic and phonetic patterns both within and across languages for the same speakers. Furthermore, MELI integrates linguistic behaviour with sociolinguistic perspectives by linking each speaker's speech data to their explicitly expressed language ideologies, elicited through the interview content. This connection allows for a more comprehensive understanding of bilingualism that bridges acoustic evidence and speakers' self-reported views on language, standardness, and identity.

\section{Corpus design and creation}
The following section provides a detailed description of the corpus and its creation process. All data was collected between February 2024 and April 2024 in person at the Speech in Context Lab at the University of British Columbia. Transcription and forced-alignment of the interview data started in May 2024 and completed in October 2024.  

\subsection{Recruitment}
Participants were recruited through a variety of methods including posts on the UBC Psychology Paid Studies List, posts on the UBC Graduate Student Community forum, posts on social media, printed flyers, and word of mouth. The recruitment criteria include the following: (1) be living in Metro Vancouver at the time of recording; (2) be born and raised in mainland China\footnote{This requirement was not further specified and relied on participants' own interpretation.}; (3) have taken either the TOEFL or IELTS exam\footnote{Both TOEFL and IELTS are language proficiency exams commonly required for admission to English-speaking universities. This requirement was included to approximate a comparable level of English proficiency.}; and (4) have attended, or be attending, a university in an English-speaking country at the time of recording. 

Before participating in the study, all participants were asked to fill out an eligibility survey. Only those who met all criteria were contacted. The eligibility survey was in both simplified Chinese and English and assumed literacy in reading both. One participant who did not read simplified Chinese was excluded from the corpus and subsequent follow-up studies. All participants who completed the interview were compensated with \$20 CAD. 

\subsection{Participants}
The corpus consists of speech of 51 Mandarin-English bilinguals (26 women and 25 men). Participants were coded with the following structure: self-identified gender (F or M), last two digits of their year of birth, and a unique letter to differentiate between participants with same identified gender and age. For example, F00A represents a female-identifying participant who were born in the year 2000. 

\begin{figure}[h]
  \begin{center}
      \includegraphics[width=\columnwidth]{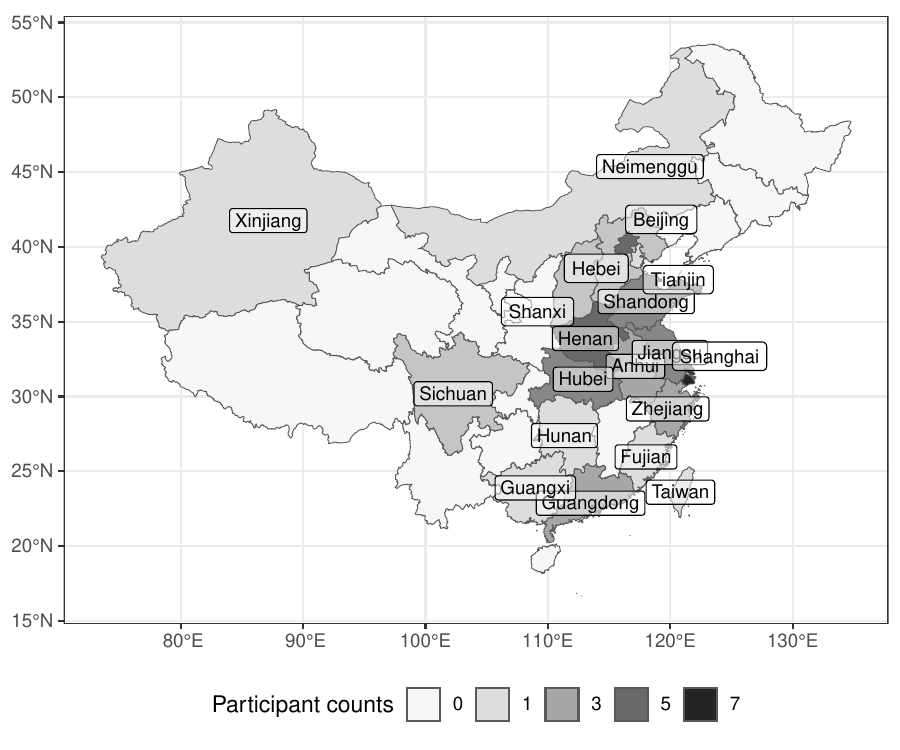}
      \caption{Geographical distribution of MELI participnats.}
      \label{fig:participants_geo}
  \end{center}
\end{figure}

\begin{figure}[h]
  \begin{center}
      \includegraphics[width=\columnwidth]{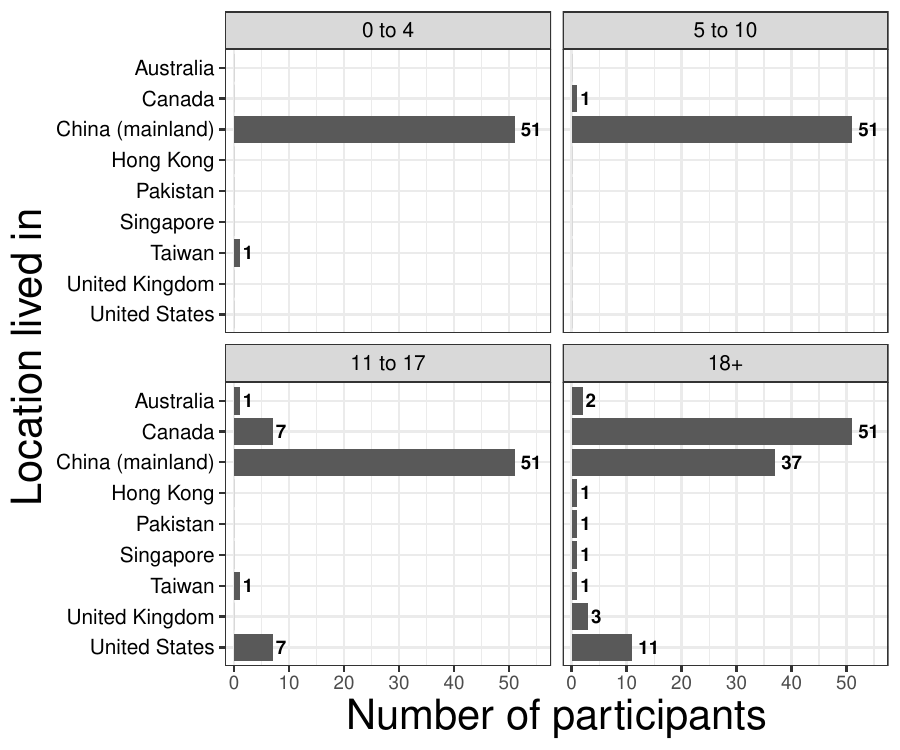}
      \caption{Residential history of MELI participants.}
      \label{fig:participants_res}
  \end{center}
\end{figure}

One goal of the MELI corpus is to include speakers from different Mandarin-speaking regions. Figure~\ref{fig:participants_geo} presents the geographical distribution of participants by province, based on their responses to the interview question ``你是哪里人?'' (\textit{nǐ shì nǎlǐ rén}). This question captures both participants' hometown (``Where are you from?'') and their regional identification (``Which region do you identify with?''). Figure \ref{fig:participants_res} provides an overview of the locations where MELI participants resided for at least 12 consecutive months across different age ranges. If participant moved from mainland China to Canada at the age of 3, both locations would be counted in the 0 to 4 panel. Overall, all participants were born and raised in mainland China and later gained residential experience in an English-speaking country, primarily Canada.

Language background is an important component of the corpus and not necessarily reflected through geographical or residential background. To begin with, bilingualism should not be treated as a homogeneous category, as bilingual speakers range from simultaneous to late learners. This is reflected in the MELI participants. The data presented here are taken directly from the language background questionnaire completed by participants, where participants rated their speaking, writing, understanding, and reading abilities for each language on a 7-point scale. A score of 1 was described as ``very low, almost no ability'' (很差), 4 as ``average'' (平均水平), and 7 as ``very high, comparable to a native speaker'' (很高, 基本相当于母语者水平). For Mandarin, the average age of acquisition was 1.23 years (range: 1--6). Mean self-ratings were 6.86 for speaking (range: 4--7), 6.70 for writing (range: 4--7), 6.88 for understanding (range: 4--7), and 6.90 for reading (range: 5--7). For English, the average age of acquisition was 6.17 years (range: 1--13). Mean self-ratings were 5.45 for speaking (range: 3--7), 5.39 for writing (range: 3--7), 5.78 for understanding (range: 3--7), and 5.96 for reading (range: 4--7). Together, these results show that MELI participants include both simultaneous and late bilinguals, with native or near-native proficiency in Mandarin and generally high proficiency in English.

\begin{table*}[!ht]
\scriptsize
\centering
\begin{tabularx}{\textwidth}{|c|c|c|C|c|C|C|c|c|}
\hline
\rowcolor{headgray}
\textbf{No} & \textbf{ID} & \textbf{YoB} &
\textbf{Interview Order} & \textbf{Gender} &
\textbf{Mandarin Lists} & \textbf{English Lists} &
\textbf{AoA (Mandarin)} & \textbf{AoA (English)}\\
\hline
1  & F00A & 2000 & M $\rightarrow$ E & F & 2, 6  & 2, 6  & 1 & 8  \\
2  & F01A & 2001 & E $\rightarrow$ M & F & 2, 9  & 2, 9  & 3 & 8  \\
3  & F01B & 2001 & E $\rightarrow$ M & F & 3, 10 & 3, 10 & 1 & 4  \\
4  & F02A & 2002 & M $\rightarrow$ E & F & 3, 9  & 3, 9  & 1 & 3  \\
5  & F02B & 2002 & M $\rightarrow$ E & F & 4, 5  & 4, 5  & 1 & 6  \\
6  & F02C & 2002 & E $\rightarrow$ M & F & 4, 6  & 4, 6  & 1 & 4  \\
7  & F02D & 2002 & M $\rightarrow$ E & F & 4, 7  & 4, 7  & 1 & 4  \\
8  & F02E & 2002 & E $\rightarrow$ M & F & 5, 7  & 5, 7  & 2 & 5  \\
9  & F03A & 2003 & E $\rightarrow$ M & F & 2, 5  & 2, 5  & 1 & 5  \\
10 & F03B & 2003 & M $\rightarrow$ E & F & 2, 10 & 2, 10 & 1 & 6  \\
11 & F03C & 2003 & E $\rightarrow$ M & F & 3, 6  & 3, 6  & 1 & 4  \\
12 & F05A & 2005 & E $\rightarrow$ M & F & 6, 9  & 6, 9  & 1 & 8  \\
13 & F89A & 1989 & M $\rightarrow$ E & F & 1, 2  & 1, 2  & 1 & 8  \\
14 & F92A & 1992 & M $\rightarrow$ E & F & 5, 10 & 5, 10 & 1 & 6  \\
15 & F92B & 1992 & M $\rightarrow$ E & F & 6, 8  & 6, 8  & 1 & 10 \\
16 & F94A & 1994 & E $\rightarrow$ M & F & 1, 3  & 1, 3  & 1 & 6  \\
17 & F94B & 1994 & E $\rightarrow$ M & F & 1, 9  & 1, 9  & 1 & 4  \\
18 & F94C & 1994 & M $\rightarrow$ E & F & 5, 8  & 5, 8  & 1 & 6  \\
19 & F94D & 1994 & E $\rightarrow$ M & F & 5, 9  & 5, 9  & 1 & 8  \\
20 & F95A & 1995 & E $\rightarrow$ M & F & 1, 7  & 1, 7  & 1 & 6  \\
21 & F95B & 1995 & M $\rightarrow$ E & F & 5, 6  & 5, 6  & 1 & 7  \\
22 & F96A & 1996 & M $\rightarrow$ E & F & 1, 10 & 1, 10 & 1 & 12 \\
23 & F97A & 1997 & E $\rightarrow$ M & F & 3, 4  & 3, 4  & 1 & 9  \\
24 & F98A & 1998 & E $\rightarrow$ M & F & 3, 8  & 3, 8  & 1 & 3  \\
25 & F99A & 1999 & E $\rightarrow$ M & F & 2, 3  & 2, 3  & 1 & 9  \\
26 & F99B & 1999 & E $\rightarrow$ M & F & 4, 10 & 4, 10 & 1 & 5  \\
27 & M00A & 2000 & M $\rightarrow$ E & M & 3, 5  & 3, 5  & 1 & 6  \\
28 & M00B & 2000 & E $\rightarrow$ M & M & 5, 9  & 5, 9  & 1 & 6  \\
29 & M01A & 2001 & E $\rightarrow$ M & M & 1, 5  & 1, 5  & 1 & 13 \\
30 & M01B & 2001 & M $\rightarrow$ E & M & 4, 9  & 4, 9  & 1 & 5  \\
31 & M02A & 2002 & M $\rightarrow$ E & M & 3, 7  & 3, 7  & 1 & 7  \\
32 & M03A & 2003 & M $\rightarrow$ E & M & 1, 8  & 1, 8  & 1 & 4  \\
33 & M03B & 2003 & M $\rightarrow$ E & M & 5, 8  & 5, 8  & 1 & 4  \\
34 & M04A & 2004 & E $\rightarrow$ M & M & 6, 7  & 6, 7  & 1 & 7  \\
35 & M04B & 2004 & E $\rightarrow$ M & M & 7, 8  & 7, 8  & 1 & 3  \\
36 & M04C & 2004 & M $\rightarrow$ E & M & 9, 10 & 9, 10 & 1 & 1  \\
37 & M05A & 2005 & M $\rightarrow$ E & M & 5, 6  & 5, 6  & 1 & 8  \\
38 & M05B & 2005 & E $\rightarrow$ M & M & 5, 7  & 5, 7  & 1 & 6  \\
39 & M89A & 1989 & M $\rightarrow$ E & M & 2, 8  & 2, 8  & 1 & 11 \\
40 & M89B & 1989 & E $\rightarrow$ M & M & 8, 10 & 8, 10 & 1 & 6  \\
41 & M94A & 1994 & M $\rightarrow$ E & M & 5, 10 & 5, 10 & 5 & 7  \\
42 & M95A & 1995 & E $\rightarrow$ M & M & 2, 7  & 2, 7  & 1 & 6  \\
43 & M96A & 1996 & M $\rightarrow$ E & M & 7, 9  & 7, 9  & 6 & 5  \\
44 & M96B & 1996 & M $\rightarrow$ E & M & 5, 8  & 5, 8  & 1 & 10 \\
45 & M97A & 1997 & M $\rightarrow$ E & M & 1, 4  & 1, 4  & 1 & 5  \\
46 & M97B & 1997 & E $\rightarrow$ M & M & 4, 10 & 4, 10 & 1 & 6  \\
47 & M98A & 1998 & M $\rightarrow$ E & M & 2, 4  & 2, 4  & 1 & 6  \\
48 & M98B & 1998 & M $\rightarrow$ E & M & 6, 10 & 6, 10 & 1 & 7  \\
49 & M99A & 1999 & M $\rightarrow$ E & M & 1, 6  & 1, 6  & 1 & 6  \\
50 & M99B & 1999 & E $\rightarrow$ M & M & 7, 10 & 7, 10 & 1 & 1  \\
51 & M99C & 1999 & M $\rightarrow$ E & M & 8, 9  & 8, 9  & 1 & 3  \\
\hline
\end{tabularx}
\caption{Basic information for each participant's interviews in the MELI Corpus.
M $\rightarrow$ E = Mandarin interview first; E $\rightarrow$ M = English interview first.}
\label{tab:ch2_metadata}
\end{table*}

\subsection{Recording setup}
The recording sessions were conducted in a quiet room at the Speech in Context Lab at the University of British Columbia. The participant and interviewer were seated across a table from each other. Both were equipped with SHURE WH20XLR headworn dynamic microphones positioned approximately 3 cm from the corner of the mouth. The microphones were connected to separate channels of a Focusrite Scarlett USBPre2 portable audio interface. Recordings were made in stereo at a 44.1 kHz sampling rate with 16-bit resolution using Audacity \citep{audacity2018} on a PC computer.

\subsection{Recording procedure}
After signing the consent form, the audio release form and completing the language background questionnaire, participants were given a quick walk-through of the procedure of the interview. This walk-through verbally (1) went over the general procedure of the recording session, (2) reassured participants that they did not need to answer questions they felt uncomfortable with, and they could withdraw from the study at any point without any consequences, and (3) they could redact any part of the interview before the release of the corpus. Participants then completed two recording sessions in Mandarin and English in one sitting. The order of the languages was counterbalanced (see Table \ref{tab:ch2_metadata} for details). 

Each recording session began with a sentence-reading task, followed by a 20--30 minute interview. Participants first familiarized themselves with the sentences before reading them aloud. Both sessions were recorded in a single audio file for each participant. In the finalized corpus, the audio file was segmented by language session.

\subsubsection{Sentence reading}

The reading materials for the sentence reading task were selected to act as stimuli for a setence-in-noise transcription task in the first author's dissertation work, which will not be discussed in the current paper. The Mandarin sentences were taken from the Mandarin Speech Perception (MSP) sentence test material \citep{fu2011}, which contains 10 lists. All lists were designed to include a comprehensive set of phonemes in the respective language. The English sentences were 10 lists taken from the English Hearing In Noise Test (HINT) \citep{nilsson1994}, which was developed for evaluation of speech perception threshold in noise. The original HINT includes 25 lists. Prior to recording, a norming survey was conducted in February 2024. Fourteen Mandarin--English bilingual raters evaluated the naturalness of sentences from all 20 HINT lists (each rater evaluated 13 lists to reduce fatigue) and from the 10 MSP lists on a continuous scale from 1 to 7. The 10 English HINT lists with the highest normalised naturalness ratings were selected for recording (Mean = 0.87, range: 0.83--0.90)\footnote{The English sentence lists used were 2, 14, 15, 18, 19, 20, 21, 22, 23, 24 \citep{nilsson1994}.}. These values are comparable to those of the Mandarin MSP lists (Mean = 0.87, range: 0.82--0.93). Each list contains 10 sentences.

In each language session, participants read two lists of sentences drawn from the selected HINT lists (English session) or MSP lists (Mandarin session). After an initial read-through, participants were instructed to repeat each sentence\footnote{The first five participants (F89A, F94A, M97A, M01A, M99A) read each sentence only once, as this repetition procedure was implemented after the initial few sessions.}. This resulted in 40 sentences per language (2 lists $\times$ 10 sentences $\times$ 2 repetitions).

\subsubsection{Interview}
Interviews were always conducted after sentence reading. The Mandarin interview focused on two main topics: (1) eliciting participant's attitude towards Mandarin varieties, Chinese languages, and Standard Mandarin with relation to their language backgrounds and (2) reflecting on participant's impression of their voice. Inspired by \citet{preston1982drawmap}, a draw-a-map task was conducted during the discussion of the Mandarin and Chinese language landscape, where participants were asked to circle the geographical regions that speak the ``most standard'' and ``least standard'' Mandarin. The English interview focused on (1) participant's experience learning English and their attitude towards different English varieties and (2) their reflection on how their voices differ across languages, if at all. A full list of sample questions are provided in the corpus documentation.

No instruction or comment on participant's speaking style was given during this session. In particular, no instruction on code-switching or specific language variety was given. 

\section{Annotation}

This section outlines the procedures for processing and annotating the recorded interviews. The pipeline follows these steps: (1) extraction of participant's channel and file segmentation by language, (2) initial transcripts using Whisper text-to-speech \citep{radford2023}, (3) hand-correction of orthographic transcripts by bilingual research assistants, (4) word and phone level forced alignment using Montreal Forced Aligner \citep{mcauliffe17mfa}, and (5) anonymization.

\subsection{Whisper speech-to-text transcription} 
The Whisper model \citep{radford2023} was used to transcribe both English and Mandarin interviews. Whisper is a pre-trained, Transformer-based sequence-to-sequence model trained for either speech recognition or a combination of speech recognition and speech translation. For this corpus, the multilingual \texttt{whisper-large-v3}\footnote{\href{https://huggingface.co/openai/whisper-large-v3}{https://huggingface.co/openai/whisper-large-v3}} model was selected, as it was the most current and suitable model for transcribing multilingual data at the time.  

Interview recordings were first segmented by language in PRAAT \citep{boersma2022}. Mandarin and English interviews were processed separately. For each language, the participant's audio channel was extracted as a WAV file and submitted to the Whisper model. A customized Python script, adapted from the \texttt{whisper-large-v3} documentation, was used to automate the transcription process. The model was configured for transcription (rather than translation) and implemented a chunked long-form algorithm with a 30-second segment length and a batch size of 16. This procedure divided long audio files into 30-second segments, transcribed each segment individually, and then merged the outputs at segment boundaries. The resulting transcriptions, including utterance-level timestamps, were exported in CSV format.

\subsection{Transcription correction and adjustment}

The auto-generated transcriptions were then hand-corrected by research assistants. The CSV files were converted to TextGrid using ELAN \citep{elan2024} for hand correction. Each TextGrid file contains four tiers: (1) task (sentence reading vs. interview), (2) automatic transcription by Whisper, (3) corrected transcription and (4) notes. Research assistants revised the Whisper-generated transcriptions in the corrected transcription tier and added notes in the notes tier. Notes included flagged identifying information and content participants requested to be redacted before the corpus release. The following conventions were followed during hand-correction:

\subsubsection{General conventions}
\begin{itemize}
  \item Unintelligible speech was transcribed as ``xxx''.  
  \item Punctuation was generally avoided, except for question marks (``?'') and possessives (marked with `` ' '').  
  \item Speech fragments were annotated using ``\&'' followed by English orthography or Pinyin (e.g., ``...\&con confident...'').  
  \item Non-speech sounds were transcribed using the labels listed in Table \ref{tab:filler-label}.  
  \item Language or speech variation such as code-switching or deviation from a participant's baseline accent was marked with ``@'' followed by a description (e.g., @e for code-switches to English, @YantaiHua for switches to the local Chinese variety spoken in Yantai). A complete list of language varieties participants switched to can be found in the corpus documentation.  
  \item Numbers were spelled out in full (e.g., ``a hundred'' or ``一百'' \textit{yì bǎi} for 100).  
\end{itemize}

\begin{table}[!ht]
\small
\centering
\begin{tabularx}{\columnwidth}{|l|X|}
\hline
\rowcolor[HTML]{BFBFBF}
\textbf{Symbol} & \textbf{Meaning} \\
\hline
\{interviewer\} & When interviewer is talking \\ \hline
\{laughter\} & Laughter \\ \hline
\{sil\} & Silence \\ \hline
\{cough\} & Coughing \\ \hline
\{chupse\} & Sucking air in between teeth \\ \hline
\{inhale\} & Inhale \\ \hline
\{click\} & Clicking \\ \hline
\{sniff\} & Sniff \\ \hline
\{exhale\} & Exhale \\ \hline
\{chchch\} & Ch-ch-ch (sound for looking for things) \\ \hline
\{micnoise\} & Microphone noise \\ \hline
\{swoosh\} & Swoosh sound \\ \hline
\{rhythm\} & Demonstrating rhythm, accent, or intonation \\
\hline
\end{tabularx}
\caption{Labels for non-speech or filler items.}
\label{tab:filler-label}
\end{table}

\subsubsection{English-specific conventions}
\begin{itemize}
  \item Filler words were transcribed using one of the following standardised forms that best matched the pronunciation: \textit{hm, uhhuh, mmhm, um, uh, mm, huh, ah, em, nn, eh, ih}. 
  \item Capitalisation was used for country names (e.g., China), brand names (e.g., Apple), languages (e.g., English, Mandarin), first-person subject (I), single letters when spelled out (e.g., A-Z), and segmented syllables (e.g., A PPLE).
\end{itemize}

\subsubsection{Mandarin-specific conventions}
\begin{itemize}
  \item The default character for the third-person pronoun was standardised to ``她'' (\textit{tā}). 
  \item Sentence-final particles and exclamations were transcribed using one of the following standardised forms that best matched participant's production: 啊~\textit{a}, 吗~\textit{ma (question)}, 呀~\textit{ya}, 吧~\textit{ba}, 呢~\textit{ne}, 咯~\textit{lo}, 嘛~\textit{ma (statement)}, 哎~\textit{ai}, 啦~\textit{la}, 嘿~\textit{hei}, 哇~\textit{wa}, 哦~\textit{o}, 哼~\textit{heng}, 滴~\textit{di}, 嘟~\textit{du}, 耶~\textit{ye}, 哈~\textit{ha}, 呐~\textit{na}, 呗~\textit{bei}, 嘞~\textit{lei}, 咦~\textit{yi}. 
  \item Common filler words were transcribed with the following: 嗯~\textit{en}, 呃~\textit{~e}, 嗯哼~\textit{en heng}, 昂~\textit{ang}, 哎哟~\textit{ai yo}.
  %\item Ambiguous syllables were transcribed using Pinyin. 
\end{itemize}

\subsection{Forced alignment}
Forced alignment was generated using Montreal Forced Aligner (MFA) v3.0 \citep{mcauliffe17mfa}, based on the hand-corrected transcriptions. This process required an audio file, an orthographic transcription, an acoustic model, and a pronunciation dictionary. The outputs were word-level and phone-level alignments for the interview audio files. 

For forced alignment of English interviews, the \texttt{English (US) ARPA acoustic model v3.0.0}\footnote{\href{https://mfa-models.readthedocs.io/en/latest/acoustic/English/English\%20\%28US\%29\%20ARPA\%20acoustic\%20model\%20v3_0_0.html}{MFA English (US) ARPA acoustic model v3.0.0.}} was employed. The dictionary, derived from \texttt{the English (US) ARPA dictionary v3.0.0}\footnote{\href{https://mfa-models.readthedocs.io/en/latest/dictionary/English/English\%20\%28US\%29\%20MFA\%20dictionary\%20v3_0_0.html}{MFA English (US) MFA dictionary v3.0.0.}}, was customized by manually transcribing out-of-vocabulary (OOV) lexical items detected during validation. These transcriptions adhered to the ARPA convention and were incorporated into the existing dictionary.  

For forced alignment of Mandarin interviews, word boundaries were first introduced using the \texttt{jieba} Python library \footnote{\href{https://github.com/fxsjy/jieba}{https://github.com/fxsjy/jieba}} since they were not marked in the original transcriptions. Forced alignment utilized the \texttt{Mandarin (China) MFA acoustic model v3.0.0}\footnote{\href{https://mfa-models.readthedocs.io/en/latest/acoustic/Mandarin/Mandarin\%20MFA\%20acoustic\%20model\%20v3_0_0.html\#Mandarin\%20MFA\%20acoustic\%20model\%20v3_0_0}{MFA Mandarin MFA acoustic model v3.0.0}} and the \texttt{Mandarin MFA dictionary v3.0.0}\footnote{\href{https://mfa-models.readthedocs.io/en/latest/dictionary/Mandarin/Mandarin\%20\%28China\%29\%20MFA\%20dictionary\%20v3_0_0.html}{MFA Mandarin (China) MFA dictionary v3.0.0}}, which provides pronunciations in International Phonetic Alphabet (IPA). OOV lexical items were identified during validation, transcribed manually using the same IPA conventions, and added to the dictionary.  

\begin{figure*}[!htbp]
  \begin{center}
      \includegraphics[width=\textwidth]{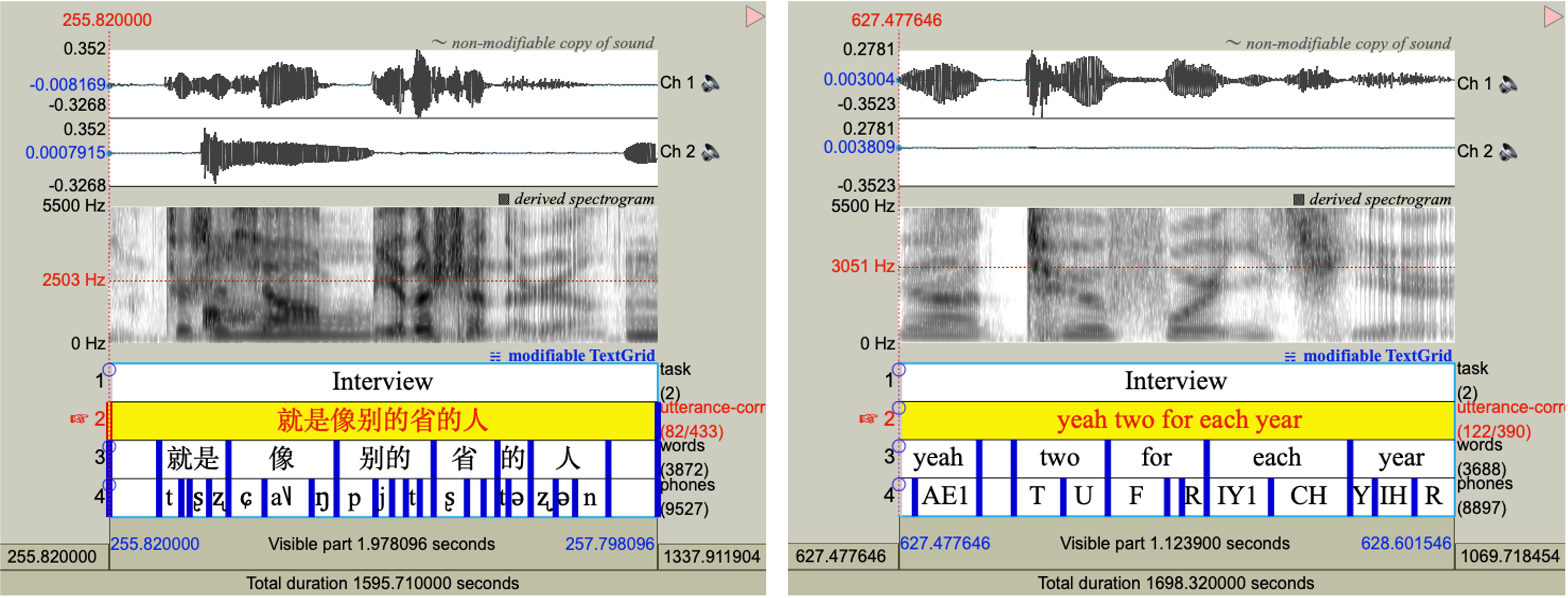}
      \caption{Examples of the forced-aligned interviews (left: Mandarin; right: English).}
      \label{fig:textgrids}
  \end{center}
\end{figure*}

The output TextGrid file includes four tiers: (1) task (sentence reading vs. interview), (2) corrected transcription, (3) word tier, and (4) phone tier. Speech fragments, unintelligible speech, and code-switched are transcribed as ``spn'' in the phone tier by MFA. Figure \ref{fig:textgrids} shows examples of the forced-aligned interviews for both languages.  

\subsection{Anonymization}

Transcribed interviews were returned to participants, who were invited to indicate any portions they wished to have redacted. These segments were marked in the notes tier by two research assistants, incorporating both participant requests made during the review process and any redaction requests expressed during the interview. In addition to participant-specified content, all personal or identifying information (e.g., names of individuals or schools) was redacted, even if not explicitly flagged. All redactions were independently identified and verified by two research assistants. Once the sections requiring redaction were identified, the corresponding audio in both the participant and interviewer channels was muted and replaced with a 1000~Hz tone using ELAN \citep{elan2024}. The associated intervals in the TextGrid files were replaced with ``[REDACTED]'' using Praat \citep{boersma2022}.

\section{Descriptive Statistics}
This section summarises the basic characteristics of the Mandarin and English interview recordings, including total duration, speaker-level variation, and lexical content in each language component of the MELI Corpus.

\subsection{Mandarin Interviews}
The Mandarin interviews include 14.7 hours of speech data including both the sentence reading task and the interview. The mean duration of speech data across participants is 17.3 minutes (range: 9.1 minutes -- 28.9 minutes). These data exclude silences in the participants' speech.  

A total of 5,799 word types (i.e. unique words) and 153,782 word tokens were produced in the Mandarin interview sessions. The total value of word types and word tokens vary across participants with a mean of 586 for word types (range: 334--923) and 3,015 for word tokens (range: 1,304--5,670). This estimation is based on the default dictionary in the \texttt{jieba} Python library used for word segmentation and takes all words into consideration regardless of language. This includes all code-switched items, unintelligible speech, and speech fragments. Figure~\ref{fig:ch2_CS}A shows a break-down of the types of tokens produced in the Mandarin interview sessions. All participants except for F03C code-switched to English during their Mandarin interview. 

\subsection{English Interviews}
Same as the Mandarin interviews, the duration of English interviews include both sentence reading task and the interview, excluding silences. The English interviews include 15.1 hours of speech data with a mean duration of 17.8 minutes across participants (range: 7.5 minutes -- 29.22 minutes).  

\begin{figure*}[htbp]
  \centering
  \includegraphics[width=\textwidth]{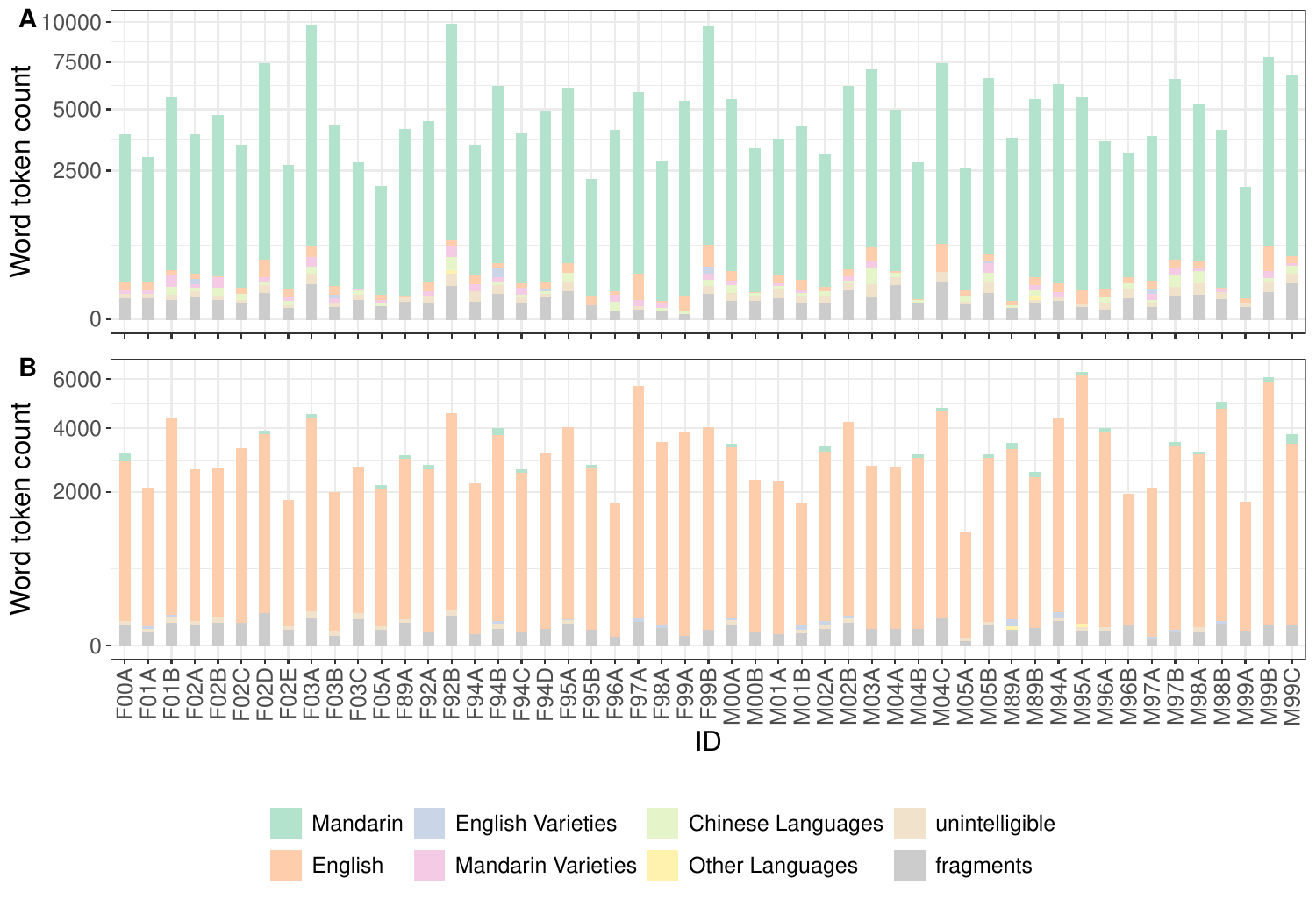}
  \caption{Distribution of code-switching in Mandarin (A) and English (B) interviews.}
  \label{fig:ch2_CS}
\end{figure*}

A total number of 4,543 word types and 129,647 word tokens were produced in the English interview sessions, with a mean word type of 516 (range: 334 to 760) and a mean word token of 2,542 (range: 955 to 5,158) across participants. Figure \ref{fig:ch2_CS}B shows a break-down of the word tokens. 23 out of 51 participants switched to Mandarin during their English interview and 19 out of 51 participants did not code switch at all. In comparison to the Mandarin interviews, MELI participants code switched more from Mandarin to English than from English to Mandarin.

\section{MELI corpus release}

Following anonymization, the MELI corpus is scheduled to be released in March 2026 through UBC Research Data Collection via Scholars Portal Dataverse under a Creative Commons Attribution 4.0 International License\footnote{\href{https://creativecommons.org/licenses/by/4.0/}{https://creativecommons.org/licenses/by/4.0/}}, which allows use for non-commercial research and educational purposes with attribution. The initial release includes anonymized audio files, corresponding transcriptions and forced alignments in Praat TextGrid format, scanned copies of maps from the draw-a-map task in the Mandarin interviews, a detailed language background and metadata summary, and a README file documenting the corpus structure. All releases can be found in the online documentation.\footnote{\href{https://meli-corpus.readthedocs.io/}{https://meli-corpus.readthedocs.io/}} Future updates will include extended annotations and alignments of the interviewer's speech. Users will be notified of updates through the repository page. 

\section{Discussion and conclusion}
Languages cannot be studied without acknowledging the variations embedded in its speakers, though fully contextualized language analysis can be challenging due to the lack of resources. The \textbf{M}andarin-\textbf{E}nglish \textbf{L}anguage \textbf{I}nterview (MELI) Corpus introduced here attempts to provide more resources for this gap, allowing studies of (1) regional varieties of Mandarin, (2) second-language accents of English, and (3) cross-language dynamics within bilingual speech. The unique format of high-quality interview recordings in MELI allow approaches both qualitatively and quantitatively, bridging together digital humanities, phonetics, and computational linguistics, and returns the voice to the people who are most intimately connected to the communities and its languages. 

\section{Acknowledgements}
The creation of the MELI corpus was approved by the University of British Columbia Behavioural Research Ethics Board (H23-03205), and was supported by the UBC Arts Graduate Research Award to the first author and SSHRC grant to the second author. We thank the MELI participants for sharing their time, voice and insights, and many members of the Speech-in-Context lab for their contribution in the creation process, especially Angelina Yuan, Dlorah Lyne Agama, Jeff Li and Sarah Ong.

\bibliographystyle{lrec2026-natbib}
\section{Bibliographical References}
\bibliography{lrec2026-example}\label{lr:ref}
\bibliographystylelanguageresource{lrec2026-natbib}
\bibliographylanguageresource{languageresource}

\end{CJK*}
\end{document}